**Recommended Citation**

Wendy, S. T. W., How, B. C., & Atoum, I. (2014). Using Latent Semantic Analysis to Identify Quality in Use ( QU ) Indicators from User Reviews. In *The International Conference on Artificial Intelligence and Pattern Recognition (AIPR2014)* (pp. 143–151). Asia Pacific University of Technology & Innovation (APU), Kuala Lumpur, Malaysia.SDIWC Publications.

```
@inproceedings{article2014,
address = {Asia Pacific University of Technology \& Innovation (APU), Kuala Lumpur, Malaysia.},
author = {Wendy, Syn Tan Wei and How, Bong Chih and Atoum, Issa},
booktitle = {The International Conference on Artificial Intelligence and Pattern Recognition
    (AIPR2014)},
isbn = {9781941968024},
keywords = {data mining,information system,intelligent,latent semantic analysis,quality in
    use,reviews},
pages = {143--151},
publisher = {SDIWC Publications},
title = {{Using Latent Semantic Analysis to Identify Quality in Use ( QU ) Indicators from User
    Reviews}},
year = {2014}
}
```

# Using Latent Semantic Analysis to Identify Quality in Use (QU) Indicators from User Reviews

Wendy Tan Wei Syn, Bong Chih How, Issa Atoum
Faculty of Computer Science and Information Technology, University Malaysia Sarawak
94300 Kota Samarahan, Sarawak, Malaysia
wendytws@siswa.unimas.my, chbong@fit.unimas.my, Issa.Atoum@gmail.com

## ABSTRACT

The paper describes a novel approach to categorize users' reviews according to the three *Quality in Use* (QU) indicators defined in ISO: effectiveness, efficiency and freedom from risk. With the tremendous amount of reviews published each day, there is a need to automatically summarize user reviews to inform us if any of the software able to meet requirement of a company according to the quality requirements. We implemented the method of Latent Semantic Analysis (LSA) and its subspace to predict QU indicators. We build a reduced dimensionality universal semantic space from Information System journals and Amazon reviews. Next, we projected set of indicators' measurement scales into the universal semantic space and represent them as subspace. In the subspace, we can map similar measurement scales to the unseen reviews and predict the QU indicators. Our preliminary study able to obtain the average of F-measure, 0.3627.

## 1 INTRODUCTION

There are various ways human could express their feelings and emotions: speech, text, gesture, facial expression and so on. Due to the advance in Internet technology, they write reviews at online websites such as Amazon and CNet after they used certain products and services. Thus, Internet becomes a medium in communicating with other people of similar interest. Reviews give huge impacts on business and social field. For business, a business company able to know how end users feel for their products from social perspective: the reviews actually can influence other people's decisions and opinions regarding certain product.

The *Quality in Use* (QU) model by International Standard Organization (ISO) is the validated measured method to measure the quality of software or products which related to the outcome after user interacted with the products [2]. The indicators for QU model are effectiveness, efficiency and freedom from risk, satisfaction and context coverage [2]. According to Atoum and Bong, the reason that drives users to use a product is if the product is perceived able to achieve particular goals such as effectiveness and efficiency [1]. In this study, we intend to automatically reveal the indicators from a pool of software reviews without human intervention. These indicators are important as they are the key decision factors to be taken into a consideration before a company purchase and adopt particular software.

As the Internet and social media thrive, the vigorous rate of increasing number of user generated reviews or related text have made the analyzing task become tedious and difficult. In addition to the incongruent context such as emotion and personal preference, the process is further complicated with the different interpretation of ambiguous word expressed by different individuals.

In this paper, we investigate a computational way to help in assimilating the mountainous reviews into useful information to allow one to make decision. We get QU indicators' related behavioral variables and use their respective measurement scales to indicate the existence of QU indicators in reviews. We assume that when we able to relate similar measurement items to the reviews, we can infer their QU indicators.

## 2 PROBLEMS

Users write reviews to express their feelings and communicate with others after they used a certain product. Most of the text mining methods on user reviews are based on feature to determine the polarity (positive, neutral or negative). There are other information that can be revealed from the reviews such as the QU indicators, which determined if ICT products able to help to achieve an organization's goals. Furthermore, interpreting reviews by human can cause bias and confusion as each of us have our own distinct interpretations. Thus, interpreting reviews sometime are subjectively depended on individual context. With this, we desire an approach to detect QU indicator from reviews without introducing bias.

## 3 RESEARCH GOAL

This study is to propose an approach to detect the three QU indicators expressed by users in reviews: effectiveness, efficiency and freedom from risk of the products. These are the characteristics from international standard to measure the quality of product.

## 4 BACKGROUND

Personal satisfaction, success of business and human safety are depended on high quality software and system [2]. It is important that product quality characteristics can be measured based on validated measurement methods according to International Standard derived from ISO/IEC 9126:1991. This international standard defines a QU model that composed of five characteristics that about "the outcome of interaction when a product is used in a particular context of use" [2]. QU model is a model that represent "the degree to which a product or system can be used by specific users to meet their needs to achieve specific goals" [2]. Examples of the goals are *effectiveness*, *efficiency*, *freedom from risk* and *satisfaction*.

For the sake of brevity, in this study, we are focusing on three characteristics: *effectiveness*, *efficiency* and *freedom from risk* as QU indicators. The definitions for the three chosen QU indicators according to ISO 9241-11 [2] are given in Table 1.

**Table 1**. Definitions of QU indicators according to [2]

| QU Indicators | Definitions |
|---|---|
| *Effectiveness* | accuracy and completeness with which users achieve specified goals |
| *Efficiency* | resources expended in relation to the accuracy and completeness with which users achieve goals |
| *Freedom from risk* | degree to which a product or system mitigates the potential risk to economic status, human life, health, or the environment |

As the formal definition of the three QU indicators from ISO document is rather brief, we reinforced the indicators with expert validated measurement scales from Human Behavioral Project [4]. The examples of the measuring scales for each indicator are shown in Table 2.

**Table 2.** Example of a portion of measurement scale used to enforce the context of *effectiveness*, *efficiency*, *freedom from risk*

| QU Indicators | Measurement scales |
|---|---|
| *Effectiveness* | -Helped set clear objectives.<br>-The overall accuracy.<br>-Number of errors. |
| *Efficiency* | -It requires a lot of time.<br>-I spent a lot of effort.<br>-The group's problem solving process was efficient. |
| *Freedom from risk* | -My decision to participate in Amazon auctions is risky.<br>-Order and structure are very important in a work environment.<br>-Will the IT work within the program use common technical resources? |

## 5 OUR METHOD

### 5.1 Latent Semantic Analysis Universal Space and Subspace

Latent semantic analysis (LSA) uses vector space model (VSM) to represent documents and words as vectors in a high dimensional vector space. The model defines the documents based on the sum of the words (or known as type) meanings. Therefore this can be used to find similar documents based on semantic meaning which does not necessitate that two similar documents must have the same words to define them as similar.

As explained by Martin and Berry [5], to create the vector space model, we form type-by-document matrix A from the documents. The rows of the matrix represent terms and column represents each document. The elements of matrix A, $a^{ij}$ are the frequencies of the particular terms ($i^{th}$) appear in each document ($j^{th}$).

Weighting functions such as log entropy, term frequency-inverse document frequency (TFIDF) usually used to normalize the documents. The purpose is to give higher weights to the terms that appear in several documents but not all and give lower weights to the terms that frequently appear in many documents which are not significant. Once we created matrix A, we convert them into orthogonal components. Orthogonal matrix contain [5]:

$$Q^T Q = I$$

Transpose of Q → ← Identity matrix
Orthogonal matrix     (5.1)

LSA used Singular Vector Decomposition (SVD) to convert matrix A into orthogonal components. SVD can represent types and documents simultaneously that able to capture the underlying semantic meaning by manipulating the number of dimensions [5]. SVD for m x n matrix A is shown as below

$$A = U \sum V^T$$

Type vectors → ← Document vectors
Diagonal matrix/ singular values      (5.2)

Type vectors and document vectors are representing words and documents coordinates respectively in the semantic space. By choosing the number of dimensions, we can find significant words and documents representations which capture the main information. In order to create k-dimensional semantic space, the SVD truncated and created $A^k$ which reduce the dimension from *r* to *k* [5]. In this way, noises are removed and it captures the important semantic structure of types of documents [5]. Types or documents that contain similar meaning are located close to each other in this k-dimensional space. Once the vector space is created, we can project query: words or documents, find similar types or documents in this high dimensional space.

We treat query as document to be projected in the space which also known as "*pseudo-document*" [5], hereby it can be represented by,

$$query = q^T U_k \Sigma_k^{-1} \quad (5.3)$$

Where $q^T$ vector is the weighted frequency of the types based on the types in query. The weights given are using weighting function. In our experiment, we build the k-dimensional semantic space (universal semantic space) using the set of paragraphs extracted from high quality journal. According to the findings presented by Martin and Berry [5], the best optimal choice for number of dimension, K is between 100 and 300. Hereby we have chosen K=300 in our experiments. Besides that, due to we are going to test on reviews, thus we also used set of Amazon reviews to build the space. We intend to build a universal semantic space that contains contexts related to both measurement scales and reviews. After we formed the semantic space, we projected our measurement scales into the space by sentences basic. We treat the measurement scales as "pseudo-document" because we want to get their positions in the semantic space.

$$measurement\ scales = measurement\ scales^T U_k \Sigma_k^{-1} \quad (5.4)$$

Measurement scales act as the contexts for each QU indicators. After we have projected all measurement scales into the semantic space, we had created a new subspace known as "scale subspace". This newly built subspace contains vectors of measurement scales. We then projected our reviews as second 'pseudo-document" into the universal space yielding reviews vectors,

$$review = review^T U_k \Sigma_k^{-1} \quad (5.5)$$

To combine both,

$$items\ subspace = U_k \Sigma_k^{-1} (reviews^T + item^T) \quad (5.6)$$

Derived from universal semantic space ⎵  Weighted type frequencies ⎵

We can see that both reviews and measurement scales are now positioned in the same universal space. They have common weighted sum of type vectors $(U_k)$ scaled by the common singular values which derived from the universal space (shown by $U_k \Sigma_k^{-1}$ in formula 5.6). We can see that both items and reviews are utilized the universal semantic space that are rich with various words usage which can bring them closer to each other based on their words meaning in a reduced dimensionality space. We use cosine similarity measure to find similar measurement scales to our testing reviews based on their vectors in the newly built "items subspace" instead of the large universal semantic space. Which means the universal semantic space are act as the platform to improve the semantic similarity between the measurement scales and reviews. To summarize, we assume that when we build universal semantic space with huge amount of related documents, it can sufficiently define different "concepts". Then the subspace is built based on our interest topics or concepts which related to the universal semantic space context.

## 6 DATA

### 6.1 Paragraph and Amazon Reviews to Build Universal Space

In order to build a universal semantic space, we adopted 95084 Amazon reviews for category "software" from SNAP project by Jure Leskovec [3]. Besides that, we also used collection of paragraphs from journals that manually extracted under Human Behavior Project by Larsen [4].

## 6.2 Measurement Scales from Human Behavior Project

To reinforce the context of the three indicators from QU: *effectiveness*, *efficiency* and *freedom from risk*, we carefully select a set of closely related behavior variables from Human Behavior Project based on the indicators' keywords (*effectiveness, efficiency, risk (*similar to *freedom from risks)*). The following table show the total number of measurement scales for each category.

**Table 3.** Number of measurement scales for each indicator

| Indicators/Constructs | Number of measurement scales |
|---|---|
| *Effectiveness* | 30 |
| *Efficiency* | 18 |
| *Freedom from risk* | 21 |

## 6.3 Annotated Reviews

In this study, we have employed a collection of 1947 annotated review sentences from the works of Atoum and Bong [1]. To the best of our knowledge there are no data sets for software quality so a gold standard is needed. The gold standard is a set of software review sentences crawled from the web and classified by the annotators. These sentences (at the end of the annotation process) will have the sentence quality-in-use topic, sentence polarity and indicating topic keyword(s). First, a set of reviews are crawled from software web sites Amazon.com and CNet.com respectively. These reviews are filtered. Then the top 10% reviews from each review rating are selected. Next, reviews are split into sentences. Finally, the sentences are given to annotators for tagging and annotation.

Before we run the experiments, we labelled each of the measurement scales and annotated reviews with the three QU indicators: *efficiency*, *effectiveness*, *freedom from risk*.

## 7 EXPERIMENTS

In this section, we explain the steps involved in predicting QU indicators from reviews.

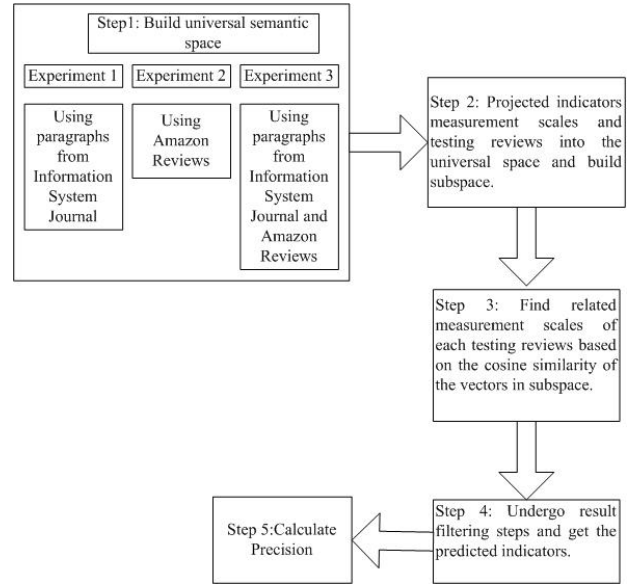

**Figure 1.** Experiment flow

Firstly, we build universal semantic space with different datasets: i) paragraphs from Information System Journal, ii) reviews, and iii) combination of (i) and (ii) above. We would like to investigate how the different universal semantic space can affect the overall performance. Next, for every universal semantic space built, we created a subspace by projecting measurement scales and testing reviews based on common term weights that have been scaled and truncated in the universal semantic space. By using this subspace, we can find similar measurement scales based on cosine similarity between the vectors presented in the subspace. We filter the returned results and get the final review's predicted indicators. We repeat each step for every testing review and report the precision.

## 8 EVALUATION

To get the predicted QU indicators, we obtain the six most similar measurement scales for each reviews sentences. Based on our study, this is the best optimum number to achieve the best accuracy. Next, we undergo several filtering steps on the results that we obtained to decide the best QU indicators for the review. The overall views of the filtering steps are illustrated by the pseudo code below:

**Table 4.** Pseudo code for filtering step

```
START
      READ scores and labelled indicators from returned LSA results
      SORT results based on highest to lowest similarity score (R1)
      READ highest score R1[0]:score and second highest score R1[1]:score from the sorted list R1
      COUNT variance = R1[0]:score – R1[1]:score
      IF variance >0.2:
          review's predicted QU indicators = R1[0]:labelled indicators
          RETURN review's predicted QU indicators
      ELSE
          For labelled indicators in R1 do
              COUNT freq= # of times labelled indicators appear in R1
          end do
          review's predicted QU indicators = max[freq]:labelled indicators
          RETURN  review's predicted QU indicators
      ENDIF
END
```

From the returned list of measurement scales, sorted by their similarity score, we need to check the variance of the similarity score between them. If the variance exceeded 0.2, we opt for the QU indicators with the highest similarity score. If the variance is less than 0.2, we infer the most similar QU indicators by taking the majority vote count of the indicators return from our result. If there is similar number of majority vote count obtained, we consider the QU indicators from the items that have the highest similarity score. Once we obtained the best predicted QU indicator for each reviews, we check with the actual labelled QU indicators and present the results in a confusion matrix. Next, we calculate precision, recall and F-measure based on the following formulas adapted from information retrieval proposed by Manning, Raghavan and Schütze [6]:

Precision, P ("Fraction of retrieved documents that are relevant")

$$\frac{TP}{TP + FP}$$

Recall, R ("Fraction of relevant documents that are retrieved")

$$\frac{TP}{TP + FN}$$

F-measure ("weighted harmonic mean of precision recall")

$$2 \times \frac{P \times R}{P + R}$$

## 9 RESULTS AND DISCUSSION

The following figure shows the F-measure in predicting the three indicators using three subspaces built from different corpus as explained in Section 6.

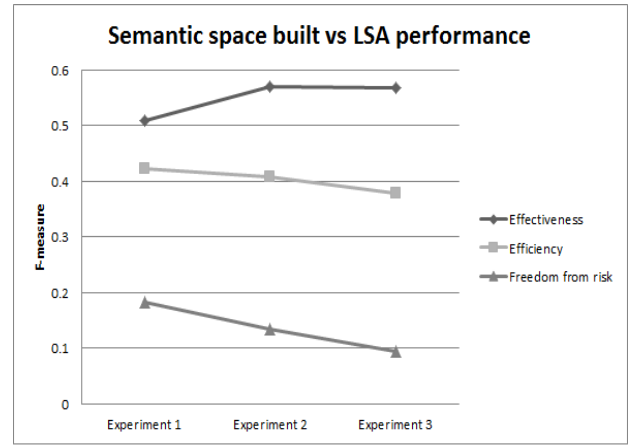

**Figure 2.** Semantic space built vs LSA performance

Figure 2 show that the overall best performance is using the journal paragraphs (Experiment 1). The overall average of F-measure is 0.3627. Although combining both journal paragraph and Amazon increases slightly the accuracy of *Effectiveness*, however, the other two indicators' accuracies declined due to the presence of noises when combined both resources. The following example illustrated correctly predicted sentences which were in agreement with experts.

Table 5. Example of review sentences predicted correctly

| Predicted Indicators | Reviews sentences |
|---|---|
| *Effectiveness* | 1. You have your choice of 3 color options: white, gray and light gray. 2. Does everything i need and more. 3. I love having everything categorized and put into a graphical form. |
| *Efficiency* | 1…..also all the programs seem to run slower than 2010. 2. Quick install and startup easy. 3. The amount of time required to complete each task. |
| *Freedom from Risk* | 1. It has some minor bugs , but nothing that can't be overcome. 2. The setup program crashed when i was installing and then it would not activate. 3. Buy at your own risk. |

We can see above that even though the items and reviews do not have much overlapping words, they still can be detected as similar. This indicates our methods able to find similar documents based on word usages. However, there are several cases where our approach was unable to detect the correct indicators.

Table 6. Example of review sentences predicted incorrectly

| Predicted indicators | Actual indicators | Review sentences |
|---|---|---|
| *Effectiveness* | *Efficiency* | i don't find any improvement in the function and only minor interface changes. |
| *Effectiveness* | *Efficiency* | better graphics quality |
| *Efficiency* | *Effectiveness* | not much different from quicken 2010! |
| *Efficiency* | *Effectiveness* | the color schemes are absolutely atrocious! |
| *Freedom from risk* | *Effectiveness* | but won't work with the browser i use firefox 3.0.4 |
| *Efficiency* | *Freedom from Risk* | it's just too buggy. |

Carefully analysis in Table 6 reveals that two sentences that have similar meaning does not necessarily end up in the same category due to different interpretation and annotation. Besides that, some of the reviews sentences are ambiguity which can be interpreted differently; in this case we need to look at the overall contexts. For example, "*product worked great*", this reviews can belongs to any indicators based on what criteria that make the author think it works great. In our experiments, the testing category will be predicted according to the similar measurement scales that we used. However, we have proved that by enriching the semantic space it can improve the LSA performance in finding related texts. It actually improved the word usage and increases the accuracy. This is based on our findings in looking into the similar items context found individually.

Based on the results, we found out that term weighting algorithm used affect the performance because some frequently occur words dominate the semantic space even though the words do not contribute much information needed. The results below show the comparison between TFIDF and log entropy.

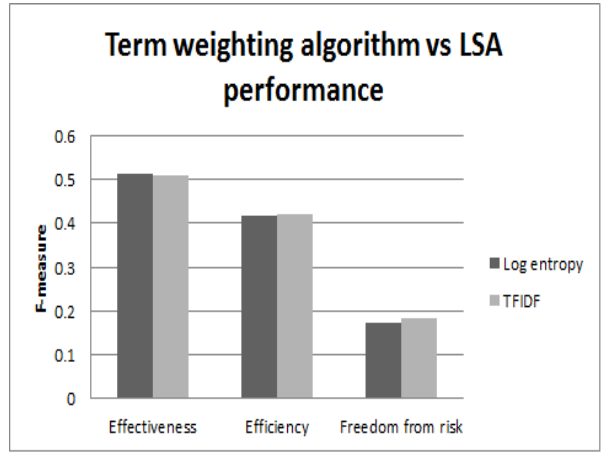

**Figure 3.** Term weighting algorithm vs LSA performance

The result showed that log entropy performs slightly better than TFIDF but still can be improved by considering other term weighting algorithm in future.

## 10 FUTURE WORKS

Although our study at this point is encouraging, the following are the planned future works.

1. We intend to obtain more measurement items to be projected into the semantic space to better detect the QU characteristics in reviews. So far, some reviews not able to be detected due to the lacking of measurement items.

2. Another approach to improve the performance is to enhance the semantic space by training it with more vocabularies. Based on the results, rich semantic space can help to improve overall performance.

3. We would need to undergo detailed preprocessing on the reviews before it analyzed by LSA due to the noise in the reviews might affect the performance.

4. In the semantic space, words with higher co-occurrence will dominate the overall information retrieval process. Log entropy performs better than TFIDF but it still cannot solve the main problem when the words are given wrong priority. We would like to take alternatives in investigating better approach in normalizing the weights to the main keywords which are more significant.

## 11 RELATED WORKS

Atoum and Bong proposed a framework in detecting QU from reviews by using topic modelling technique (Latent Dirichlet Allocation) to generate several keywords. They also used semi-supervised learning to calculate polarity of sentences [1].

Wang et. al works related to opinion mining, currently most of the methods are on feature based. Their proposed feature-based vector model contains features and also reviewer's opinion on the features. Besides that, they also consider the relationship between words and punctuation. [7]

In terms of finding related documents or document classification works, some of the well-known methods are kNN classification, Naïve bayes, vector space model, neural networks and etc.

## 12 CONCLUSION

To conclude, we had proposed an approach to detect *quality in use* (QU) characteristics from reviews by using Latent Semantic Analysis (LSA). We adapted a set of measurement items from Human Behavior Project to determine if reviews are comply to QU characteristics. We proposed the method of LSA universal semantic space and subspace. We enriched the universal semantic space with reviews and QU indicators' measurement scales context and build subspace to map similar reviews and measurement scales based on their semantic meaning. The reported results showed that QU indicators in reviews can be predicted with average F-measure of 0.3627. Several limitations caused this works even challenging where there exist different interpretations between our measurement scales and annotated reviews, ambiguous words usage. We realize that the usage of human words is varied and we shall improve the detection process by incorporate more vocabularies to enhance the semantic space and introducing better information retrieval algorithms to rigorously capture the main context.

## 13. ACKNOWLEDGEMENT

We would like to thank Universiti Malaysia Sarawak and MOHE who funded this project through grant ERGS/ICT07(01)/1018/2013(15). We also like to express our sincere gratitude to Kai Larsen, the Director of Human Behavior Project at University of Colorado Boulder for allowing us to use the behavior variables in the study.